\documentclass[a4paper,twoside]{article}

\usepackage{cite}
\usepackage{amsmath,amssymb}%,amsfonts
\usepackage{algorithmic}
\usepackage{graphicx}
\usepackage{CJKutf8}
\usepackage{textcomp}
\usepackage{xcolor}
\usepackage{times}
\usepackage{soul}
\usepackage{url}
\usepackage[utf8]{inputenc}
\usepackage[small]{caption}
\usepackage{multicol}
\usepackage{pslatex}
\usepackage{apalike}
\usepackage{algorithm2e}
\usepackage[bottom]{footmisc}
\usepackage{booktabs}
\usepackage[switch]{lineno}
\usepackage{tcolorbox}
\usepackage{multirow}
\newcommand{\eq}[1]{\begin{small}\begin{align}#1\end{align}\end{small}}
\usepackage{float}
\usepackage[greek,english]{babel}
\newcommand{\ci}{~\cite} \newcommand{\re}{~\ref} 

\DeclareMathOperator*{\argmax}{arg\,max}
\newtcolorbox{mymathbox}[1][]{colback=white, sharp corners, #1}
\usepackage{SCITEPRESS}

\begin{document}
\title{Topic Modeling with Fine-tuning LLMs and Bag of Sentences\thanks{This is the preprint of the accepted journal version of the published conference paper by Johannes Schneider  titled ``Efficient and Flexible Topic Modeling using Pretrained Embeddings and Bag of Sentences'' which appeared at the International Conference on Agents and Artificial Intelligence(ICAART) in 2024.\cite{sch24eff}.}}
%\titlerunning{Topic Modeling using Pretrained Embeddings and BoS}
\author{\authorname{Johannes Schneider}%\sup{1} and Michalis Vlachos\sup{2}
\affiliation{ University of Liechtenstein, 9490 Vaduz, Principality of Liechtenstein}%\sup{1}
\email{johannes.schneider@uni.li}
}
% \author{
%     Johannes Schneider
%     \affiliations
%     University of Liechtenstein, Vaduz,FL
%     \emails
%     johannes.schneider@uni.li
% }
%\author{\authorname{First Author Name\sup{1}\orcidAuthor{0000-0000-0000-0000}, Second Author Name\sup{1}\orcidAuthor{0000-0000-0000-0000} and Third Author Name\sup{2}\orcidAuthor{0000-0000-0000-0000}}
%\affiliation{\sup{1}Institute of Problem Solving, XYZ University, My Street, MyTown, MyCountry}
%\affiliation{\sup{2}Department of Computing, Main University, MySecondTown, MyCountry}
%\email{\{f\_author, s\_author\}@ips.xyz.edu, t\_author@dc.mu.edu} }

%\maketitle
%\begin{abstract}
%Extension for JOurnal
%We have introduced a novel algorithm called FT-Topic used for fine-tuning sentence transformer for topic modeling. Specifically, the part on data generation is a significant advancement of the state-of-the-art, ie., the final model outperforms prior models. This change is reflected in a completely new section including pseudocode and a new figure. In turn, also the evaluation is enhanced to cover the novel algorithm. We also restructured the manuscript by providing a separate model section. We have also rewritten significant parts of the abstract, introduction, related work and conclusion. We have also avoided verbatim copying of large sections.

\abstract{Large language models (LLM)'s are increasingly used for topic modeling outperforming classical topic models such as LDA. Commonly, pre-trained LLM encoders such as BERT are used out-of-the-box despite the fact that fine-tuning is known to improve LLMs considerably. The challenge lies in obtaining a suitable (labeled) dataset for fine-tuning. In this paper, we use the recent idea to use bag of sentences as the elementary unit in computing topics. In turn, we derive an approach \emph{FT-Topic} to perform unsupervised fine-tuning relying primarily on two steps for constructing a training dataset in an automatic fashion. First, a heuristic method identifies pairs of sentence groups that are either assumed to be of the same or different topics. Second, we remove sentence pairs that are likely labeled incorrectly. The dataset is then used to fine-tune an encoder LLM, which can be leveraged by any topic modeling approach using embeddings. In this work, we demonstrate its effectiveness by deriving a novel state-of-the-art topic modeling method called \emph{SenClu}, which achieves fast inference through an expectation-maximization algorithm and hard assignments of sentence groups to a single topic, while giving users the possibility to encode prior knowledge on the topic-document distribution. Code is at \url{https://github.com/JohnTailor/FT-Topic}.%\emph{anonymized URL - \url{https://drive.google.com/drive/folders/1di0za6SMPNHy7A5OFaVOpEkryfY_79je?usp=sharing}}.%.
}
\onecolumn \maketitle \normalsize \setcounter{footnote}{0} \vfill

\section{\uppercase{Introduction} }
The landscape of topic modeling has evolved significantly with the advent of large language models (LLMs), which have shown superior performance over classical models like Latent Dirichlet Allocation (LDA)\ci{ble03}.  Traditional methods such as LDA, despite their foundational role in text mining and their elegant mathematical formulation, face inherent limitations, particularly in their handling of frequent words and the bag of words (BoW) assumption, which can lead to fragmented and less coherent topic assignments. Also LDA's optimization objective, i.e. minimizing perplexity, performs worse than other measures such as PMI\ci{new10}. The shortcomings of classical models has driven the search for more sophisticated techniques that better align with human perceptions of topics.

Large language models, particularly those leveraging deep learning architectures, have transformed various natural language processing (NLP) tasks. Pre-trained models like BERT offer contextual embeddings\ci{dev18} that capture intricate relationships between words in a way that static word vectors and traditional methods cannot. 
So far, for topic modeling only relatively few attempts have been undertaken to leverage contextualized vectors\ci{men22,gro22}. Existing approaches use LLMs out-of-the-box for topic modeling, despite evidence suggesting that fine-tuning can significantly enhance their performance. The main hurdle for fine-tuning LLMs lies in acquiring suitable (labeled) datasets, which are often scarce and costly to produce.

In response to this challenge, we introduce \emph{FT-Topic}, a novel approach that utilizes unsupervised fine-tuning to optimize LLMs for topic modeling. Our method is inspired by using a sequence of a few sentences as the elementary unit of analysis rather than individual words or entire documents. The underlying idea is that such a text fragment contains (in most cases) sufficient information to be assigned to a topic and it is also small enough to be assigned to typically only one or few topics.

The FT-Topic approach involves constructing a training dataset through two primary steps. First, we employ a heuristic method to identify pairs of sentence groups that are likely to belong to either the same or different topics. This initial step generates a broad set of potential training pairs. Second, we refine this set by removing pairs that are likely mislabeled, ensuring higher quality data for fine-tuning. The resulting dataset allows us to fine-tune any encoder LLM, thereby improving its embedding capabilities for topic modeling tasks.

Our approach integrates seamlessly with existing topic modeling techniques that utilize embeddings, offering enhanced performance and flexibility. We specifically demonstrate the improved effectiveness of \emph{FT-Topic} for a novel the state-of-the-art method \emph{SenClu} that also treats sequences of sentences as the elementary unit while deriving the topic model.\footnote{This is a journal version of the conference paper\cite{sch24eff}, where SenClu was introduced.} We assume that (i) such an elementary unit of sentences can be assigned to one topic and (ii) two elementary units are independent. As such, it is an analogue to the bag of words model that regards a word as an elementary unit, underlying classical models such as LDA. However, the BoW models commonly lead to the situation, where topics can change after almost every word \ci{gru07,sch18}, which is highly unnatural. Other approaches have employed document clustering\cite{gro22}, which treats the entire document as an elementary unit, which is a strong deviation from topic modeling assuming that topics can vary across the document. The bag of sentences balances granularity between these extremes, providing a more coherent basis for topic extraction. 
Another advantage of the use of a bag of sentences model is that conceptually each sentence group forming an elementary unit is likely to belong to one or only few topics. In particular, in our inference, each sequence is firmly classified into a single topic, simplifying both the computational process and the user's understanding. 

Our inference mechanism in algorithm \emph{SenClu} is inspired by expectation maximization. It utilizes the aspect model to efficiently map sentences to topics through an extension of the K-Means algorithm, which clusters sets of data points rather than individual points. This method significantly accelerates the inference process compared to more complex deep learning models\ci{men22,dien20} and variational inference\ci{mia16}, while offering greater adaptability for users at the same time. While this approach may not be as quick as methods that do not capture multiple topics within documents, it consistently delivers high-quality topic identification. Overall, our methodology achieves cutting-edge performance with reasonable computational demands, providing a versatile tool tailored to user requirements.

Our contributions are as follows:\\
\begin{enumerate}
    \item We introduce an automatic method for fine-tuning (sentence-based) LLM encoders for topic novel topic model relying on heuristic training dataset construction with a quality improvement step.     
    \item We state a novel topic model along with the bag of sentence model (BoS), which utilizes pretrained sentence embeddings. This model strikes a balance between traditional models based on individual words, known as the bag of words, and models based on entire documents, such as document clustering used for topic modeling.
    \item We present a new inference technique based on a form of "annealing" that integrates clustering with the aspect model. For deriving topics—specifically, the ranking of words within a topic from sentence to topic assignments—we introduce an effective measure that combines the frequency of a word in a topic with its relative importance within that topic, automatically filtering out infrequent and irrelevant words.
    \item By testing on various datasets and comparing with several models, our approach demonstrates modest computational requirements while achieving superior performance in terms of topic coherence and topic coverage, evaluated through a downstream task. Additionally, we conduct a comprehensive review, assessing other significant factors such as the practical utility of these methods from the perspective of end-users.   
\end{enumerate}

%(for abbreviations consult Table\ref{tab:not})

\section{\uppercase{Topic Model}} \label{sec:tomo}
Our model aligns with the well-known aspect model\ci{hof01}, which calculates topic-document and word-document probabilities. It is a generative model, i.e., it allows sampling from the obtained probability distributions to generate documents. A fundamental shift in our approach is the move away from a purely word frequency-based generative model, which has its limitations. Instead, we focus on short sequences of sentences as the smallest unit of analysis for topic assignment, i.e., Bag of Sentences (BoS). Additionally, our model can be viewed as a clustering enhanced with priors. That is, our model incorporates cluster priors, specifically topic-document probabilities.
\emph{Formal definition:}
We are given a set of documents $D$. A document $d \in D$ is a sequence of groups of sequences $d=(g_0,g_1,...,g_{n-1})$ with each group $g_i=(s_{j},s_{j+1},...,s_{j+n_s})$ being a sequence of $n_s$ consecutive sentences and $j \in [i\cdot n_s,(i+1)\cdot n_s-1]$. Furthermore, $s_j$ is the $j$-th sentence in document $d$. In turn, each sentence $s_j=(w_0,w_1,...)$ is a sequence of words.\footnote{Typically, in topic modeling words also include numbers, but not any punctuation.} The same word can occur multiple times in a sentence, i.e., it can happen that $w_i=w_j$ for $i\neq j$. Analogously, the same sentence can occur multiple times in a document.

Classical topic modeling dating back to the aspect model in 2001\ci{hof01} establishes a joint probability distribution across words and documents $p(w,t)$ to compute for each document $d \in D$ and word $w$ a probability distribution $p$ of topics for a word $p(w|t)$ and for a document $p(t|d)$. That is, it is implicitly assumed that each document can have multiple topics. Topics are seen as latent variables in a generative model. It naturally contains a distribution $p(d)$ indicating the likelihood of a document by applying basic laws of conditional probability.  We maintain the same model but use groups of sentences $g$ rather than words $w$. As in the standard aspect model we assume conditional independence of a sequence of sentences $g$ and a document $d$ given a topic $t$: 	
\eq{	
		p(g,d):= p(d)\cdot p(g|d) \label{eq:dat}\\
		p(g|d):=\sum_t p(g|t)\cdot p(t|d) \label{eq:amod}	
}
While for our topic modeling approach we rely on conditional independence as in the aspect model, in our fine-tuning process discussed next, we only rely on the partitioning of documents into sentence groups (BoS), the order of sequence groups is relevant for the training data generation process.

\section{\uppercase{FT-Topic: Fine-Tuning LLMs for Topic Modeling}}
Large language models (and priorly word vectors) are commonly leveraged to compute word (and sentence) embeddings. That is, instead of using one-hot encodings of words, continuous vector representations are used. So far, out-of-the-box word vectors such as GLOVE or pre-trained language models such as BERT generating contextualized word embeddings have been employed. While these embeddings have been successful, one might wonder whether they are really ideal for the task of topic modeling. First, they are not trained for similarity computation in the context of topic modeling. That is, BERT, for instance, trains models to predict missing words and sentence order. In topic modeling, we ideally have that two words with their contexts originating possibly from different documents are deemed to be similar if they are assigned the same topic. However, as topic assignments are unknown, this cannot easily be done. To obtain embeddings in an unsupervised manner that are similar if the embedded text belongs to the same topic, we employ the following model: \\
\noindent\emph{Key assumptions:}
We view a document as a set of groups of sentences (see Section \ref{sec:tomo}). A single group is assumed to convey meaningful information for topic assignment, though the actual assignment might also depend on the entire document. This is in contrast to classical topic modeling focusing on individual words, where many words (e.g., ``a'',``the'',``is'') cannot be assigned to a topic in a meaningful way without context.
For our data generation, we also assume stronger local semantic relationships among sentence groups: Oftentimes, adjacent groups of sentences in a document have the same topic, i.e., are similar, while groups of sentences from other documents are more likely to stem from other topics, i.e., are dissimilar. In turn, this allows us to identify training data for fine-tuning in an unsupervised manner, e.g., we can identify pair of sentence groups that should be similar (i.e., those nearby in a document) and pairs of sentence groups that should be more dissimilar (i.e., those in different documents). 
\\
\noindent\emph{Generating training data and fine-tuning loss:}
Based on our prior assumption, for each sentence group in a document the next and prior sentence groups are said to be similar. In contrast, a random group from another document is dissimilar. We chose two sentence groups as negative samples. It is easy to increase the number of training samples, e.g., by choosing more random sentence groups as negative samples or assuming that more sentence groups from the same document should be considered similar. This choice can be made dependent on the dataset. We did not experiment with these options but rather stuck to the choices described. 
At this point, we have for each sentence group positive and negative samples that can be leveraged for fine-tuning using an adequate loss function. 
Using this training data, we can fine-tune a pre-trained LLM, i.e., we rely on sentence encoders \cite{rei19}. We are only left with choosing an objective for optimization. There are a few options for losses like the contrastive loss and triplet loss \cite{sch23sur}. We use the triplet loss\cite{cha10} defined as follows:
\eq{
&\mathcal{L}(A,P,N)= &\operatorname{max} (||v_A-v_P||_2-||v_A-v_N||_{2}+m ,0) %\nonumber
}
where 
$A$ is an anchor input (i.e., a group of sentences $g$), 
$P$ is a positive input that should be similar to $A$, 
$N$ is a negative input that should be dissimilar to $A$, 
$m$ is a margin between positive and negative pairs, and 
$v_A,v_P,v_N$ are the embeddings of $A$, $P$, $N$.
That is our dataset $T$ for fine-tuning is organized as a set of triplets $T=\{(A,P,N)\}$.

\begin{figure*}[h]
  \centering
\includegraphics[width=1\linewidth]{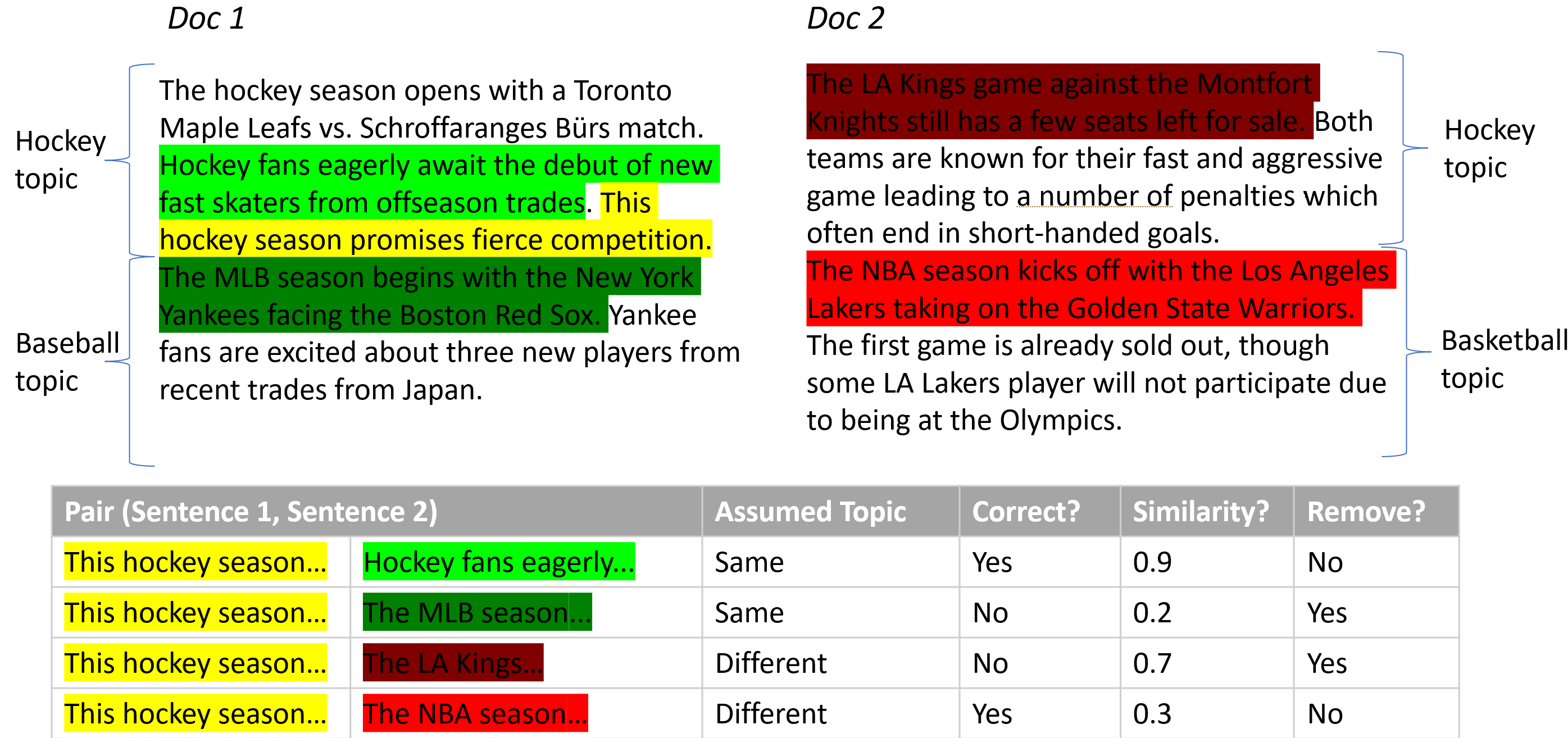}
\caption{Overview of training data generation for fine-tuning assuming a corpus $D$ of two documents and three distinct topics using single sentences. For the sentence \emph{This hockey season...} we sample sentences assumed to be in the same and distinct topic, wrong samples are removed based on similarity computation using a non-fine-tuned LLM. \label{fig:datafig}}
\end{figure*}

\noindent\emph{Improving training data quality:}
However, training data quality is generally low, when relying on our locality assumption, as it is fairly common that the assumption is violated in one of two ways: (i) group of sentences assumed to belong to the same topic belong to different topics and (ii) group of sentences assumed to belong to different topics belong to the same topic. Error (i) is common if topics frequently change within a document, which often happens if a document exhibits many topics. Error (ii) is a major concern if there are only relatively few topics. For example, if there are just four topics across all documents and each topic occurs equally often, every fourth pair assumed to be from a different topic should be assumed to be from the same topic. 
To improve training data quality, we remove training samples that (most) likely suffer from one of the two errors. Thus, we need an estimate on how likely a pair of sentences is incorrect judged as either from the same or different topic.  Non-fine-tuned models have been proven to work well for this task (though not perfect as we argue), thus we might remove samples, where the similarities computed based on embeddings of a non-fine-tuned model is indicative that the considered pair is not correct. One might make the decision which samples to keep and to remove to focus specifically to reduce either error (i) or error (ii) or both jointly. We discuss and evaluate two of these three options in the paper\footnote{We have evaluated all three, but found no benefit in outcomes for the 3rd one.}, i.e., remove pairs to reduce error (i) by removing pairs with low similarity that are assumed to be from the same topic and (ii) we remove triplets, i.e., a positive and a negative pair, if the difference of similarity of the positive pair minus that of the negative pair is small. For simplicity, we just remove a fixed fraction of all pairs that most likely suffer from an error. That is, we compute values indicating the likelihood that a training sample is incorrect based on similarity values for all samples and remove those with highest likelihood. The high level approach is shown in Figure \ref{fig:datafig}.
More precisely, we remove a fixed fraction $f_{pos}$ of triplets $(A,P,N)$ using similarity values from positive samples $(A,P)$ based on the Euclidean distance $||h(A)-h(P)||_2$. A value of 0 indicates that the vectors are identical and larger values indicating greater dissimilarity. Thus, we remove those with largest values, as in this case $A$ and $P$ are likely dissimilar although they are assumed not to be.
We also remove a fixed fraction $f_{tri}$ using similarity values combined from positive and negative samples, i.e., $||h(A)-h(P)||_2-||h(A)-h(N)||_2$. We remove again those with largest values, as in this case, fairly likely $A$ and $P$ are quite dissimilar (although assumed not be)  or $A$ and $N$ are similar or both.
The full algorithm \emph{FT-Topic} is stated in pseudocode in Algorithm \ref{alg:FT-Topic}. As any deep learning model combined with an optimization procedure, it contains a large number of parameters. We only state those that are non-standard for fine-tuning or must be explicitly set according to library we used in our implementation, i.e., the Python library called Sentence-transformer Version 3.0.1. 

\begin{algorithm}[htp]
    \caption{FT-Topic}  \label{alg:FT-Topic}
    \begin{algorithmic}[1]
        \begin{small}
            \STATE \textbf{Input: } \small{Docs $D$, Model $E$ to be fine-tuned}
            \STATE \textbf{Output: } Fine-tuned model $A'$ 
            %\STATE 
            \STATE $f_{pos}:=0.08$;  $f_{tri}:=0.24$  \COMMENT{For data cleaning: Fraction of samples to remove based on similarity values}
            \STATE $m:=0.16$ \COMMENT{Margin for the triplet loss}
            %\COMMENT{Compute similarity using positive and negative sentence groups}            
            \STATE $nNeg:=2$ \COMMENT{Number of negative sentence groups $N$ chosen for an anchor $A$ and positive sentence group $P$}            
            \STATE $ep:=4$ \COMMENT{fine-tuning epochs}
            \STATE Tokenize docs $d \in D$ into a sequence of sentences $d:=(g_0,g_1,...)$. Each sentence group $g_i=(s_0,s_1,...,s_{n_s-1})$ is disjoint and consists of $n_s$ sentences. Each sentence $s_i$ is tokenized into words.
            \STATE \COMMENT{Generate raw dataset of triplets $T=\{(A,P,N)\}$}                        
            \STATE $T=\{\}$ \COMMENT{Initial training dataset}            
            \STATE \textbf{For each } $d \in D$: 
            \STATE  \hspace{0.5em} \textbf{For } $i$ from $0$ to $|d|-1$:
            \STATE  \hspace{1em} \textbf{For } $j$ from $1$ to $nNeg$:
            \STATE \hspace{1.5em}\COMMENT{Add triplet for sentence group after $g_i \in d$}
            \STATE  \hspace{1em}  \hspace{0.5em} $g_{N}$:= Random $g \in d'$ for random $d'\in T\setminus\{d\}$          
            \STATE  \hspace{1em} \hspace{0.5em} \textbf{if} $i<|d|-1$ \textbf{then } $T:=T \cup (g_i \in d,g_{i+1} \in d,g_{N})$
            \STATE \hspace{1.5em}\COMMENT{Add triplet for sentence group before $g_i \in d$}
            \STATE  \hspace{1em} \hspace{0.5em} $g_{N}$:= Random $g \in d'$ for random $d'\in T\setminus\{d\}$
            \STATE  \hspace{1em} \hspace{0.5em} \textbf{if } $i>0$ \textbf{ then } $T:=T \cup (g_i \in d,g_{i-1} \in d,g_{N})$
            \STATE  \hspace{1em} \textbf{end}
            \STATE  \hspace{0.5em} \textbf{end}
            \STATE \textbf{end}
            \STATE \COMMENT{Remove likely incorrect triplets from $T$ using similarities based on model $E$}            
            %\STATE \textbf{For each } sample $(A,P,N) \in T$: 
            \STATE  $(v_{A},v_{P},v_{N}):=$ Embeddings of $(A,P,N)$ using model $E$
            \STATE Remove fraction $f_{pos}$ of all triples from $T$ with largest $||v_A-v_P||_2$
            \STATE Remove fraction $f_{tri}$ of all triples from $T$ with largest $||v_A-v_P||_2-||v_A-v_N||_2$
            \STATE $A'$:= Fine-tuned model $A$ on data $T$ using tripet loss with margin $m$ for $ep$ epochs
            \STATE  \hspace{0.5em} $s(v_{anc},v_{pos}$
            \STATE \textbf{return } model $A'$
        \end{small}
    \end{algorithmic}	
\end{algorithm}

\section{\uppercase{SenClu Topic Model}}
The SenClu Model leverages the BoS model introduced in Section \ref{sec:tomo}. We represent an elementary unit, i.e., a sequence of sentences, not as token (one-hot encoding) but as a continuous vector. In the same fashion, we represent topics not as one-hot encodings but as continuous vectors. From a clustering perspective, we use cluster centroids as topic vectors being the mean of points within each cluster (as in k-Means++\ci{art07}). In contrast, to simple k-means clustering, we incorporate cluster priors, specifically topic-document probabilities. 

Unlike the typical generative probability model in the aspect model (or LDA\cite{ble03}), the probability of a sentence belonging to a topic within a document in our view is represented as a binary outcome -- either 0 or 1—stemming from our cluster assignments. That is, we perform hard assignments, where a sentence group either belongs to a topic or not. We do not estimate the document probability $p(d)$, i.e., we simply use a uniform distribution as in other works, e.g.,\cite{sch18}. This is motivated by the assumption that all documents are considered roughly equally likely.  We define the remaining missing terms $p(g|t)$ and $p(t|d)$ of the aspect model (Equation \ref{eq:amod}) as follows:	
\eq{
p(g|d)&=\frac{\max_{t} \{h(g,t)\cdot p(t|d)\}}{\sum_{g} \max_{t} \{h(g,t)\cdot p(t|d)\}}  \label{eq:m1} \\
&\propto\max_{t} \{h(g,t)\cdot p(t|d)\} \nonumber\\	
h(g,t) &:= cos(v_g,v_t) \label{eq:m1a}\\
p(t|d)& := \frac{\alpha+\sum_{i<|d|} 1_{t=\argmax_{t'}\{h(g_i,t')\cdot p(t'|d)\}}}{t\cdot(1+\alpha)}\label{eq:m2}
\\ & \propto \alpha+\sum_{i<|d|} 1_{t=\argmax_{t'}\{h(g_i,t')\cdot p(t'|d)\}} \nonumber
}
%%It is important to recognize that both $p(g|d)$, representing the probability of a sentence sequence given a document, and $p(t|d)$, the probability of a topic given a document, are probability distributions as defined. 
Note that both $p(g|d)$ and $p(t|d)$ are probability distributions by definition, i.e., it is easy to verify that  $\sum_g p(g|d)=1$ and $\sum_t p(t|d)=1$. However, the normalization constants to obtain probability distributions are not relevant in our inference algorithm. The distribution $p(t|d)$ (Equation \ref{eq:m2}) states that the probability of a topic $t$ given a document is proportional to the number of sentence sequences $g_i$ assigned to $t$. We also added a smoothing constant $\alpha$ that gives a user some control over whether to prefer documents being assigned to few or many topics (similarly to the topic prior $\alpha$ in LDA). We discuss it in more depth later.
The distribution $p(g|d)$ is not derived from mere counts of word, i.e., group of sentences, occurrences. These would make limited sense, as most groups within typical corpora $D$ for topic modeling would occur only once. In our approach, we conceptualize both a topic $t$ and a sequence of consecutive sentences $g$ as vectors that encapsulate semantic features. We employ a sentence vector $v_g$ generated from a pretrained sentence transformer, specifically a (fine-tuned) sentenceBERT, used in our evaluation. For the topic vector $v_t$, we compute it by averaging the sentence vectors $v_g$ that are assigned to a specific topic $t$. The greater the similarity between the vectors $v_g$ and $v_t$, the stronger the association between the topic $t$ and the sentences in $g$. We measure this similarity using cosine similarity, which is standard for high-dimensional vectors, although the dot-product also provides very similar results. More formally, the term $p(g|d)$ is calculated based on a cosine value $h(g,t)$(Equation \ref{eq:m1a}) expressing the similarity of the group of sentence vector $v_g$ and the topic vector $v_t$ for $t$.
As a second component to compute $p(g|d)$, we also utilize $p(t|d)$ serving as prior incorporating the context of the group $g$ given by the document $d$.

%$sim$ we decided on the simplest one being the dot-product, i.e. $sim(v_s,v_t)=v_s\cdot v_t$. %The pre-trained sentence vectors $v_s$, i.e., a the sentence embedding vector $v_s$ stems from a model like sentenceBERT and $v_t$ depends on all sentences belonging to topic $t$. 
% This is the common strategy that has already been employed for non-contextual word vectors .

%We compute $p(t|d)$ dependent on semantic scores $g(s,t)$ of the words in the document $d$ (Equation \ref{eq:m3}).  We model the idea of looking for keywords in a document and aggregating their score, i.e., $g(s,t)$. The parameter $\alpha$ impacts the number of topics per document. The larger $\alpha$ the more concentrated the topic-document distribution, i.e., the fewer topics per document.	
%With the help of these equations we derive an efficient algorithm for inference.  

% and (\ref{eq:pwt})

\subsection{\uppercase{Computing Word-topic Scores}}
The standard practice in presenting topic model results is to display the most probable words from the word-topic distribution $p(w|t)$. Given the advent of LLMs and the challenge of interpreting lists of words as topics, it has also been proposed to use LLMs themselves to interpret $p(w|t)$. However, looking at most probable words also comes with advantages as this can be done quickly and using LLMs adds another lense of interpretation (as well as the risk of hallucinations). Therefore, we also explicitly compute word-topic scores. 
Since our approach involves assigning sentences to topics, yielding $p(g|t)$, a comparable method of presentation would be to display the most probable sentences. However, this method has three notable limitations. Firstly, this format is unconventional for users, as it diverges from the typical outputs of existing topic models showing word lists. Secondly, because sentences are generally long and may contain words irrelevant to the topic, this can lead to misunderstandings. Third, it requires more reading.
To derive a list of words as for the word-topic distribution $p(w|t)$ in the classical aspect model (or LDA) from $p(g|t)$, we assign each sentence group to a topic and subsequently each word within a group to the same topic. This allows us to calculate the relative frequency of each word within a topic. A limitation of this method is that frequent but non-discriminatory words like "is," "are," "a," "the," etc., tend to show high probabilities across all topics due to their ubiquitous presence in text. These words are less useful for differentiating topics. To address this, we calculate a score $score(w|t)$ for each word within a topic, based on both the frequency of the word in the topic and a measure of its relative importance to that topic compared to others. The higher this score, the more indicative the word is of the topic. The problem of ignoring such words is well known across various NLP tasks. We could therefore use standard measures such as term frequency and inverse document frequency (TF-IDF)\ci{has14}. However, we perform a more elaborated measure targeted specifically to topic modeling.

\emph{Definition of word-topic scores}: 
The frequency $n(w|t)$ represents the number of times a word $w$ is assigned to topic $t$, $n(w|d)$ denotes the occurrences of word $w$ in document $d$, and the overall frequency $n(w)$ in the corpus is calculated as $n(w)=\sum_t n(w|t)$. The frequency score is determined using a damped frequency formula: $\sqrt{\max(n(w|t)-n_{min},0)}$. Damping is a common technique in NLP to reduce the overemphasis on frequency; for example, logarithmic or square root transformations are also used in computing TF-IDF.
The term $n_{min}$ is conceptualized as the baseline frequency of words, accounting for random distribution or artificial occurrences, such as a word appearing predominantly in one document. It establishes a minimum threshold incorporating the average expected word count $n(w)/|T|$ under random uniform assignments across topics, the standard deviation $std(n(w|t))$ of word counts across topics, and the maximum occurrences $\max_d n(w|d)$ of word $w$ in any document, defined as $n_{min}:=n(w)/|T| + std(n(w|t)) + \max_d n(w|d)$. This criterion implies that for a word to be considered indicative of a topic, its occurrence should exceed the sum of its expected value for a random (uniform) distribution, its observed variability across topics, and its peak occurrences in individual documents.

In addition to the frequency measure, we assess the "relative" relevance of a word to a topic, defined as the excess probability $p(t|w)-1/|T|$. The probability $p(t|w)$ is approximated by the proportion of the word's assignments to a topic relative to its total appearances. If a word exclusively appears under one topic ($p(t|w)=1$), it is deemed highly relevant to that topic, even if infrequently used. Conversely, if a word is uniformly distributed across all topics ($p(t|w)=1/|T|$), it lacks topical significance, leading to a relative relevance of zero.

The overall score for a word in relation to a topic, $score(w|t)$, combines these frequency and relevance metrics: $\sqrt{n(w|t)-n_{min}}\cdot (p(t|w)-1/|T|)$. Words yielding zero or negative scores are not reported. Occasionally, a topic may be characterized by only a few or no words with positive scores, often occurring when a cluster is dominated by highly specialized or broadly common words, alongside a few topic-specific words that appear sporadically. For instance, there might be a cluster ``sports'' with the word sports appearing 1000 times and a second cluster with just ten appearances. In this case, both the relevance and frequency score are small for the second cluster (but high for the first). %Formulas for all terms are stated also in Algorithm \ref{alg:SenClu}.

 %The expected frequency gives the number of assignments to a topic, . Words that occur only that often for a topic should have zero score, since they are not indicative of that topic. 
%This idea has been commonly expressed in other contexts, e.g., in information retrieval `term frequency–inverse document frequency'' (Tf-idf) that combines frequency of a term with the specificity of a term, which is quantified as an inverse function of the number of documents in which it occurs\ci{jon72}. In our setting, we are less interested in the occurrence within documents but with the assignments of words to topics. That is, we weigh the frequency by the ``inverse topic frequency'' (itf). Formally,
% \eq{
% n(w|t)&:= \sum_{d \in D} \sum_{s \in A_{t,d}} \sum_{w_i \in s} |\{i| w_i=w\}|\\
% n(w)&:=\sum_t n(w|t)\\
% p(t|w)&:= \frac{n(w|t)}{\sum_{t} n(w|t)}\\
% score(w|t)&:=\sqrt{n(w|t)-n_{min}}\cdot (p(t|w)-1/|T|)\\
% n_{min}&:=n(w)/|T| + std(n(w|t))+\max_d n(w,d)
% }

%\mathrm{idf}(w, t) =  \log \frac{|\{w_i| w_i=w, w_i \in d, d \in D\}|}{|\{t|w=w_i, w_i \in s, s \in A_{t,d}, d \in D\}|}
%p(w|t)&:= \frac{\sum_{d \in D} \sum_{s \in A_{t,d}} |\{w_i| w_i=w, w_i \in s\}|}{\sum_{d \in D} \sum_{s \in A_{t,d}} |s|}
%\mathrm{idf}(w, t) =  \log \frac{|\{w_i| w_i=w, w_i \in d, d \in D\}|}{|\{t|w=w_i, w_i \in s, s \in A_{t,d}, d \in D\}|}
\begin{algorithm}[htp]
    \caption{SenClu}  \label{alg:SenClu}
    \begin{algorithmic}[1]
        \begin{small}
            \STATE \textbf{Input: } \small{Docs $D$, nTopics $k$, Prior $\alpha$}
            \STATE \textbf{Output: } Topic-document distribution $p(t|d)$, Word-topic score $score(w|t)$
            \STATE $p(t|d):= 1/k$; $T:=[1,k]$; $epochs:=10$; $c(\alpha):=\max(8,\alpha)$, $n_s:=3$ 
            \STATE Tokenize docs $d \in D$ into a sequence of sentences $d:=(g_0,g_1,...)$. Each sentence group $g_i:=(s_0,s_1,...,s_{n_s-1})$ is disjoint and consists of $n_s$ sentences. A sentence $s_i$ is tokenized into words.
            \STATE Pretrained (and fine-tuned) sentence embedder model $M:s\rightarrow v_s$ with dimension $d_s:=|v_s|$
            \STATE Initialize topic vectors $v_t \in \{s | s \in d, d \in D\}$ using k-means++ and cosine similarity             % \ci{sad20}\ci{reh11} % \STATE $p(w|t):= 1/|W|$+noise %$\forall d,t:  \forall w,t:  %\COMMENT{Add randomness}        % \STATE $ n(w,t):=0$  %\forall w,t:%\COMMENT{Assignments of word $w$ to topic $t$; Initialize with prior}
            \FOR{$i \in [1,epochs]$}            
            \STATE $A_{t,d}=\{\}$ \COMMENT{$\forall t \in T,d \in D$} \COMMENT{Begin of E-step}
            \FOR{$d \in D$}
            \STATE $r:=$ random number in $[0,1]$
            \STATE $i:= 1 \text{ if } r<0.5+i/(2\cdot epochs) \text{ else } 2$
            \FOR{$g \in d$}            
            \STATE $t_{g,d} = \argmax^i_{t} \{cos(v_g,v_t)\cdot p(t|d)\}$ \COMMENT{$\argmax^i$ gives the argument that yields the $i$-th largest value, i.e., 1 gives the largest}
            \STATE $p(t|g,d) =  1  \text{ if } t_{g,d}=t \text{ else } 0$
            \STATE $A_{t_{g,d},d}:=A_{t_{g,d},d} \cup g$            
            \ENDFOR
            \ENDFOR            
            \STATE $v_t:=\frac{\sum_{d \in D} \sum_{s \in A_{t,d}} v_s}{\sum_{d \in D} |A_{t,d}|}$ \COMMENT{Begin of M-step}
            \STATE $p(t|d):= \frac{|A_{t,d}|+c}{|d|+c}$
            \STATE $c(\alpha)=\max(c(\alpha)/2,\alpha)$            
            \ENDFOR                    
            \STATE \emph{Compute word-topic scores }
        \STATE $n(w|t)= \sum_{d \in D} \sum_{g \in A_{t,d}} \sum_{s \in g} \sum_{w_i \in s} 1_{w_i=w}$
        %\STATE $score(w|t)=\frac{n(w|t)^{1.5}}{\sum_{t} n(w|t)}$
        \STATE  $n_{min}:=n(w)/|T| + std(n(w|t))+\max_d n(w,d)$
        \STATE  $p(t|w):= \frac{n(w|t)}{\sum_{t} n(w|t)}$        
        \STATE  $score(w|t):=\sqrt{n(w|t)-n_{min}}\cdot (p(t|w)-1/|T|)$        
        \STATE \textbf{return } $p(t|d)$, $score(w|t)$
        \end{small}
    \end{algorithmic}	
\end{algorithm}
% known (Glivenko-Cantelli theorem) that the empirical distribution converges to the true distribution and

\subsection{\uppercase{Inference}} \label{sec:inf}
To estimate model parameters, we maximize the likelihood of the data represented as $\prod_d \prod_{g \in d} p(g,d)$, following our model definitions (Equations $\ref{eq:dat}$, $\ref{eq:m1}$, and $\ref{eq:m2}$). Traditional inference methods such as Gibbs sampling or variational inference are slow. Enhancing these methods, for example by collapsing variables in a Gibbs sampler, is challenging. Instead, we employ expectation-maximization (EM) combined with clustering concepts to expedite the inference process. Mirroring the approach in \cite{sch18}, we devise an EM algorithm based on standard probabilistic reasoning using frequencies of sentence-topic assignments. The EM algorithm includes two main steps: the E-step and the M-step. During the E-step, we estimate latent variables, specifically the probability $p(t|g,d)$ of a topic given sentences $g$ in a document $d$. In the M-step, we keep the topic distribution $p(t|g,d)$ fixed and aim to maximize the loss function with respect to the parameters.
We implement hard assignments in our model, assuming that a sentence in a document correlates to exactly one topic with a probability of one, while the probability for all other topics is zero. Thus, the topic $t_{g,d}$ of a sentence group $g$ in a document $d$ is determined to be the most probable topic. Formally: %, i.e., adjusting Equation (\ref{eq:m2}) accordingly we get:			
    \eq{		
    &t_{g,d} := \argmax_{t} \{h(g,t)\cdot p(t|d)\} \label{eq:ass} \\
    &p(t|g,d) = \begin{cases}  \label{eq:siass}
    1   \hspace{1em} t_{g,d}=t\\
    0  \hspace{1em} t_{g,d}\neq t\\
    \end{cases}	\\
    &A_{t',d}:=\{g|t_{g,d}=t', g \in d\}%\sum_{t} \sum_{g \in d} p(t|g,d)}     %t_{g,d}
    }

The term $A_{t,d}$ represents the groups of sentences $g$ that are assigned to topic $t$ in document $d$. Our approach, outlined in Equation \ref{eq:ass} and subsequent equations, deviates from models like PLSA and LDA, where every word within a document is typically assigned a probability distribution across all topics. Our method offers computational benefits but also presents challenges, such as the tendency to get trapped in local minima. To address this, we may employ an annealing technique that introduces a decreasing level of randomness to help escape these local optima.
In the M-Step, our goal is to optimize the parameters. Similar to the approach described in Equations (9.30) and (9.31) in \ci{bis06}, we define the function $Q(\Theta,\Theta^{old})$ which represents the complete data log likelihood dependent on the parameters $\Theta$:

\eq{
&\Theta^{new} = \argmax_{\Theta} Q(\Theta,\Theta^{old}) \\ &\text{ with } 	Q(\Theta,\Theta^{old}):= \sum_{d,t}  p(t|D,\Theta^{old}) \log  p(D,t|\Theta)  \label{eq:opt}		
}
		
The optimization problem described in Equation (\ref{eq:opt}) may be approached through various methods, such as using Lagrange multipliers. Unfortunately, straightforward analytical solutions are impractical due to the complexity outlined in the model's Equations (\ref{eq:m1} and subsequent equations). However, we can reconsider the method of inferring parameters $p(g|t)$ and $p(t|d)$ from a different perspective. Suppose we are provided with all sentence groups $A_{t,d}$ assigned to topic $t$ in document $d$ across a collection of documents $D$. In this case, we define the topic vector $v_t$ as simply the mean of these sentences:

\eq{
v_t:=\frac{\sum_{d \in D} \sum_{g \in A_{t,d}} v_g}{\sum_{d \in D} |A_{t,d}|}
}

To determine the remaining parameters, our inference utilizes a frequentist approach, deriving the empirical distribution: The probability of a topic given a document is determined by the proportion of the document’s sentences assigned to that topic. Under reasonable assumptions, this maximum likelihood estimate corresponds to the empirical distribution, as explained and justified in section 9.2.2 in \ci{bar12}:

\eq{
p(t|d)&:=\prod_{i<|d|} p(g_i|d) \text{ (Using Equation \ref{eq:m2})} \approx \frac{|A_{t,d}|}{|d|} 
}
We employ a value $c(\alpha):=c'\cdot\alpha$ for a user-specified value $\alpha$ and a value $c'$ decayed throughout optimization. Finally, we obtain
\eq{
p(t|d)\approx \frac{|A_{t,d}|+c(\alpha)}{|d|+k\cdot c(\alpha)}
}

\begin{table*}[htp]%[b]{0.46\textwidth}
    %\vspace{-10pt}
    \vspace{15pt} %5
    \caption{Datasets. Classes are human defined categories. The Gutenberg dataset contains books in multiple languages}\label{tab:ds}    
    \centering
    %\begin{scriptsize}
    \begin{footnotesize}    %\setlength\tabcolsep{1pt}
        \begin{tabular}{ l  c  c  c  c}
            \toprule
         Dataset & Docs & \#Words/doc & Vocabulary & Classes \\	
         \midrule
            New York Times (NYT) & 31,997 & 690 & 25,903 & 10\\ 
            20Newsgroups&18,625 & 122 &37,150 & 20\\ 
            Gutenberg& 9,417 & 1,768 & 290,215 & 50 \\
            Yelp & 29,820 &191 & 75,791& 1 \\
            \bottomrule
        \end{tabular}   %\end{scriptsize}
    \end{footnotesize}
   % \vspace{-6pt}
\end{table*} 

\begin{table*}%[htp!]
  %\vspace{-6pt}
		\caption{The 20 Newsgroups} \label{tab:top}
		\scriptsize
		\setlength\tabcolsep{3pt}
		\begin{tabular}{| p{\textwidth}  |}\hline 
			comp.graphics, comp.os.ms-windows.misc, comp.sys.ibm.pc.hardware, comp.sys.mac.hardware, comp.windows.x, rec.autos, rec.motorcycles, rec.sport.baseball, rec.sport.hockey, sci.crypt, sci.electronics, sci.med, sci.space, misc.forsale,talk.politics.misc,talk.politics.guns, talk.politics.mideast, talk.religion.misc, alt.atheism, soc.religion.christian \\ \hline
		\end{tabular}	
 %\vspace{-10pt}
\end{table*}
The parameter $c(\alpha)$ serves two primary functions. Initially, it aims to enhance the optimization process through its gradual decrease, preventing issues like local minima and poor initial setups. The parameter $c$ starts at an initial high value, $c_0:=8$, and decreases each epoch $i$ to $c_i:=\max(c_{i-1}/2, \alpha)$, where $\alpha$ is a user-defined minimum value to be detailed later. This decay process helps to minimize the impact of early, potentially suboptimal topic-document distributions by maintaining a non-zero probability $p(t|d)$ for each topic in the early phases of optimization. This precaution prevents premature convergence on poorly defined topics, particularly for short documents where a few sentences might be erroneously assigned to an underdeveloped topic. Initially, $p(t|d)$ starts as a uniform distribution, but without intervention, it could quickly polarize, assigning most or all sentences to a single topic and reducing the probabilities of other topics to zero or near-zero. Thus, the assignment to a not yet well-formed topic $t$ is likely not changed in later epochs (Equation \ref{eq:ass}), and some topics might not be considered any more since their probability has become zero. Thus, conceptually, early in the optimization, we do not yet fully trust the resulting topic to document assignments since they are subject to change. 
In turn, we smoothen the resulting probability $p(t|d)$. One might also view the decaying process as a form of (simulated) annealing since it fosters changes of topic sentence assignments initially but slowly makes them less likely by making the distribution $p(t|d)$ more concentrated.

The second function of $c(\alpha)$ relates to its final value $\alpha$. It is to determine the diversity of topic coverage in the final sentence assignments within a document. A higher $\alpha$ value promotes a more even, dispersed topic-document distribution, akin to the role of the hyperparameter $\alpha$ in LDA, which influences the dispersion across topic-document distributions. The choice of $\alpha$ depends on the document length and user preference, indicating hypothetically how many sentence groups should be assigned to each topic, with typical values ranging from 0 to the average number of sentence groups per document.

Initialization involves randomly selecting sentence group vectors $v_s$ to define initial topic vectors $v_t$, similar to the k-means++ strategy. This method is susceptible to local minima, particularly if an outlier influences the initial topic vector, which could misrepresent the cluster's central tendency. To counteract this, we occasionally reassign sentence groups from one cluster to a neighboring cluster. Initially, a document might be assigned to its second most likely cluster with a probability of about 0.5, a probability that we gradually reduce to zero within half the epochs to allow for eventual convergence.

Our method called SenClu is stated in Algorithm \ref{alg:SenClu}. It processes a corpus $D$, given the number of desired topics $k$ and a prior $\alpha\geq 0$ originating from user preference for topic diversity per document. The exact initial setting of $c(\alpha)$ is not crucial as long as it remains above 2. We suggest that a topic should span a few sentences, thus a grouping size from 1 to 5 sentences is deemed optimal, smoothing the influence of rare words or sentences that don't easily align with a specific topic without additional context. A very large group size, approximating the average number of sentences per document, would treat topics as singular units, affecting the model's granularity.

%than the value $c$ given by the user and decay it to the user

\section{\uppercase{Evaluation}}

We conducted both qualitative and quantitative assessments using four benchmark datasets, four methods, and three metrics. Additionally, we evaluated the influence of parameters like the margin $m$ for the triplet loss and the fraction $f_{pos}$ and $f_{tri}$ of removed data  to improve data quality for \emph{FT-Topic} as well as the number of topics $k$, the number of sentences $n_s$ (per group), and the prior $\alpha$ for \emph{SenClu}.\\
\noindent\textbf{Settings:} Our experiments were executed on an Ubuntu 20.4 system equipped with Python 3.9 and Pytorch 1.13, running on a server with 64 GB of RAM, a 16-core AMD Threadripper 2950X CPU, and an NVIDIA RTX TI 2080 GPU. Unless otherwise noted, our settings included $k=50$ topics, $\alpha=2$, and the top 10 words from each topic. We performed three trials for each setup and report both the average and the standard deviation. Code is at \url{https://github.com/JohnTailor/FT-Topic}.\\
\noindent\textbf{Methods:} As a foundational comparison, we utilized LDA\ci{ble03} implemented via Python's Gensim 4.3\ci{reh11}, which is used in most papers as a baseline and, thus, allows indirect comparisons across many models. Furthermore, we state results for BERTopic\ci{gro22} and TopClus\ci{men22}, which employ fixed pre-trained contextualized embeddings and clustering techniques. These methods are discussed in the related work section and represent state-of-the-art models using similar methodology.

\noindent\textbf{Datasets and Pre-processing:}
For \emph{FT-Topic} and \emph{SenClu}, we tokenize documents into sentences using a straightforward rule-based tokenizer\ci{sad20}. These sentences are then converted into contextual sentence embeddings using a sentence encoder model. As a base model for fine-tuning we use sentence transformers\ci{rei19}. 
For LDA, we utilize the default tokenizer provided by Gensim, whereas for BERTopic and TopClus, the preprocessing is integrated within the respective libraries. As a post-processing step for all methods, we lemmatize topical words and remove duplicates prior to selecting the top 10 words for analysis. The datasets listed in Table\re{tab:ds} have been previously employed in topic modeling studies\ci{sch18,men22}, with the exception of the Gutenberg dataset, which includes books in various categories and languages from the public Gutenberg library. The NYT dataset is annotated with two human categorizations into 10 classes, and we utilized pre-defined countries for location-based categorizations.

%            Embedding   Multiple topics per Doc  Dimensionality Reduction Method complexity  Clustering Separate word-top extraction
% BerTopic   Document    N                          Y                          Density based, HDBScan   Y
% TopClus    Word       Y                           Y                           With custom loss        N
% OUrs       Sentence   Y                           N                           With custom loss        Y  

\begin{table}%[htp]	
\vspace{-5pt}
\caption{Quantitative comparison for fractions $f_{tri}$ and $f_{pos}$  for FT-Topic (using SenClu as topic model). }\label{tab:resmarg}		
    \centering	%\begin{footnotesize}%\setlength\tabcolsep{2pt}    
		\begin{tabular} {| c | l | l  | l | }
			\hline			
			Dataset & $f_{tri}, f_{pos}$ & NMI  & PMI  \\ \hline
\multirow{4}{*}{20News} & 0, 0 & 0.47\tiny{\text{$\pm$} 0.009}&0.71\tiny{\text{$\pm$} 0.034}\\
& 0.32, 0 & 0.48\tiny{\text{$\pm$} 0.007}&0.79\tiny{\text{$\pm$} 0.015}\\
&0.24, 0.08 & 0.48\tiny{\text{$\pm$} 0.005}&0.8\tiny{\text{$\pm$} 0.016}\\
 &0, 0.32 &   0.48\tiny{\text{$\pm$} 0.006}&0.81\tiny{\text{$\pm$} 0.01}\\ \hline 
\multirow{4}{*}{Guten} & 0, 0 &  0.23\tiny{\text{$\pm$} 0.007}&0.62\tiny{\text{$\pm$} 0.075}\\
& 0.32, 0 & 0.25\tiny{\text{$\pm$} 0.01}&0.68\tiny{\text{$\pm$} 0.069}\\
&0.24, 0.08 & 0.26\tiny{\text{$\pm$} 0.005}&0.68\tiny{\text{$\pm$} 0.041}\\
 &0, 0.32 &   0.27\tiny{\text{$\pm$} 0.007}&0.68\tiny{\text{$\pm$} 0.071}\\ \hline 
\multirow{4}{*}{NYT} & 0, 0 &  0.29\tiny{\text{$\pm$} 0.034}&0.65\tiny{\text{$\pm$} 0.043}\\
&0.32, 0 & 0.31\tiny{\text{$\pm$} 0.028}&0.8\tiny{\text{$\pm$} 0.044}\\
&0.24, 0.08 & 0.31\tiny{\text{$\pm$} 0.039}&0.77\tiny{\text{$\pm$} 0.068}\\
 &0, 0.32 & 0.33\tiny{\text{$\pm$} 0.01}&0.76\tiny{\text{$\pm$} 0.052}\\ \hline 
\multirow{4}{*}{Yelp} & 0, 0 & -  & 0.51\tiny{\text{$\pm$} 0.024} \\
&0.32, 0 & - & .6\tiny{\text{$\pm$} 0.019}\\
&0.24, 0.08 &- & 0.6\tiny{\text{$\pm$} 0.024}\\
 &0, 0.32 &- & 0.56\tiny{\text{$\pm$} 0.015}\\  \hline 
		\end{tabular}  
	%\end{footnotesize}	
	%\vspace{-6pt}
\end{table}

\begin{table}%[htp]	
\vspace{-5pt}
\caption{Quantitative comparison for margin $m$ for FT-Topic (using SenClu as topic model). }\label{tab:resf}	
    \centering	%\begin{footnotesize}%\setlength\tabcolsep{2pt}    
		\begin{tabular} {| c | l | l  | l | }
			\hline			
			Dataset & Margin \text{$m$}& NMI  & PMI  \\ \hline
\multirow{3}{*}{20News} & 0.08 &0.47\tiny{\text{$\pm$} 0.01}&0.78\tiny{\text{$\pm$} 0.026} \\
&0.16 & 0.48\tiny{\text{$\pm$} 0.005}&0.8\tiny{\text{$\pm$} 0.016} \\ 
 & 0.32 & 0.49\tiny{\text{$\pm$}0.01} & 0.81\tiny{\text{$\pm$}0.02} \\\hline 
\multirow{3}{*}{Guten} & 0.08 &  0.26\tiny{\text{$\pm$} 0.007}&0.74\tiny{\text{$\pm$} 0.035}\\
&0.16 & 0.26\tiny{\text{$\pm$} 0.005}&0.68\tiny{\text{$\pm$} 0.041}\\
 & 0.32 &0.25\tiny{\text{$\pm$} 0.005}&0.56\tiny{\text{$\pm$} 0.058}\\\hline
\multirow{3}{*}{NYT} & 0.08 & 0.32\tiny{\text{$\pm$} 0.021}&0.7\tiny{\text{$\pm$} 0.064}\\ 
&0.16 & 0.31\tiny{\text{$\pm$} 0.039}&0.77\tiny{\text{$\pm$} 0.068}\\
 & 0.32 & 0.32\tiny{\text{$\pm$} 0.037}&0.8\tiny{\text{$\pm$} 0.03}\\\hline
\multirow{3}{*}{Yelp} & 0.08 & - &0.61\tiny{\text{$\pm$} 0.025}\\
&0.16 & - &0.6\tiny{\text{$\pm$} 0.024}\\
 & 0.32 & -&0.56\tiny{\text{$\pm$} 0.023}\\\hline
		\end{tabular}  
	%\end{footnotesize}	
	%\vspace{-6pt}
\end{table}

\subsection{Quantitative Evaluation}
We concentrated on evaluating topic coherence and topic coverage. Coherent topics are typically more meaningful and sensible. For assessing topic coherence, we calculated the normalized Pointwise Mutual Information (PMI) score\ci{new10} at the document level as defined in \ci{sch18}, using the English Wikipedia dump dated 2022/10/01 as an external reference. The PMI score is favored over measures like perplexity as it aligns more closely with human judgment\ci{new10}.

For topic coverage, we employed a downstream task, specifically clustering based on topic models, which we then compared against predefined human categories as detailed in \ci{men22}. This comparison is quantified in the Normalized Mutual Information (NMI) score.

We also considered computation time as a key metric, acknowledging the importance of energy consumption during a climate crisis. That is, we also aim at reducing computation stated as one design principle for green data mining \cite{schn23re}. Topic models are utilized by a wide range of researchers and practitioners for whom computation time is crucial. The reported computation time covers both the training and inference phases for topics across a corpus, excluding the time spent computing evaluation metrics like PMI, as this is consistent across all methods. Included in the computation time are steps specific to each topic model, such as tokenization, computing embeddings, and all processes necessary to derive word-topic scores and topic-document distributions. 

\smallskip

\noindent \textbf{Results FT-Topic:} 
Results for the sensitivity analysis of the (hyper)parameters are shown in Tables \ref{tab:resmarg} and \ref{tab:resf}. 
Table \ref{tab:resf} also shows the benefit of performing data filtering, i.e., cleaning using similarity computation based on a non-fine tuned LLM. When we use data filtering ($f_{tri}>0$ or $f_{pos}>0$ or both), the NMI and PMI increase considerably. For PMI, gains are even more noticeable. When it comes to the question how to filter data, i.e., using similarity estimates based on both positive and negative samples ($f_{tri}$) or just positive samples ($f_{pos}$), we see that it makes some but limited differences. The differences also seem to be dataset dependent. Focusing more on removing positive samples seems to help the NMI but its impact on the PMI depends on the dataset. For the margin $m$ we also observe dataset dependent behavior. 
Overall, we observe that there is limited sensitivity to the exact choice of the hyperparameters, which means that our algorithm is easy to use without much adjustments to default parameters. 

\smallskip

\noindent \textbf{Results SenClu:} 
Results for the sensitivity analysis of the (hyper)parameters are shown in Tables \ref{tab:resk}, \ref{tab:resns}, and \ref{tab:resalpha}. These parameters have some, though limited impact, indicating that at least on a quantitative level, the algorithm behaves fairly insensitively to the parameters.\\%number of topics $k$  has little impact on the metrics. Both larger $\alpha$ and larger $n_s$ improve metrics somewhat. Larger $\alpha$ implies that documents tend to have more topics, as such similar sentence (sequences) from different documents can be grouped more easily.  

Table \ref{tab:res} displays the results of our method comparison. The quantitative analysis reveals that SenClu (w/o FT-Topic) and TopClus perform best in generating relevant topics. Surprisingly, SenClu without topic modeling surpasses TopClus in topic coverage on two of the three datasets examined. SenClu with topic modeling surpasses TopClus on all datasets for all metrics; notably, Yelp does not provide clustering data. This suggests that despite TopClus being specifically designed for clustering and assessed on similar tasks, our approach demonstrates that extensive clustering optimization and dimensionality reduction may not be necessary for effective topic modeling. In fact, these techniques could potentially hinder performance if the underlying assumptions, like the noisiness of embeddings, are unmet. Our approach also excels in PMI calculations, indicating significantly higher topic coherence. This is further evident when analyzing the actual topic words, where other models often include generic, irrelevant terms. One notable drawback of TopClus is its lengthy computation time, taking several hours for even moderately sized datasets. In contrast, SenClu is considerably quicker, requiring only a few minutes, although still slower than LDA and BerTopic, which typically finish in about a minute. FT-Topic requires also a considerable amount of computation, but it is still much faster than TopClus while outperforming it. However, relying solely on quantitative metrics may not fully capture the effectiveness of the models, as detailed in our overall and qualitative evaluations in Section \ref{sec:over} and the qualitative evaluation discussed next.

\begin{table}%[htp]	
\vspace{-5pt}
\setlength\tabcolsep{1pt}
		\caption{Quantitative comparison for number of topics $k$ for SenClu. }\label{tab:resk}
    \centering	%\begin{footnotesize}%\setlength\tabcolsep{2pt}    
		\begin{tabular} {| c | l | l  | l | }
			\hline			
			Dataset & nTopics $k$ & NMI  & PMI  \\ \hline
\multirow{3}{*}{20News} & 25 & 0.46\tiny{\text{$\pm$}0.01} & 0.8\tiny{\text{$\pm$}0.02} \\
&50 & 0.47\tiny{\text{$\pm$}.003} & 0.79\tiny{\text{$\pm$}.037} \\ 
 & 100 & 0.47\tiny{\text{$\pm$}0.0} & 0.73\tiny{\text{$\pm$}0.02} \\\hline
\multirow{3}{*}{Guten} & 25 & 0.2\tiny{\text{$\pm$}0.01} & 0.83\tiny{\text{$\pm$}0.07} \\
&50 & 0.2\tiny{\text{$\pm$}.003} & 0.67\tiny{\text{$\pm$}.03} \\ 
 & 100 & 0.2\tiny{\text{$\pm$}0.01} & 0.75\tiny{\text{$\pm$}0.03} \\\hline
\multirow{3}{*}{NYT} & 25 & 0.29\tiny{\text{$\pm$}0.01} & 0.77\tiny{\text{$\pm$}0.03} \\
&50 & 0.28\tiny{\text{$\pm$}.021} & 0.78\tiny{\text{$\pm$}.025} \\ 
 & 100 & 0.28\tiny{\text{$\pm$}0.01} & 0.75\tiny{\text{$\pm$}0.05} \\\hline
\multirow{3}{*}{Yelp} & 25 & - & 0.65\tiny{\text{$\pm$}0.03} \\
&50 & - & 0.62\tiny{\text{$\pm$}.007}\\ 
 & 100 & - & 0.6\tiny{\text{$\pm$}0.01} \\\hline
		\end{tabular}  
	%\end{footnotesize}	
	%\vspace{-6pt}
\end{table}

\begin{table}%[htp]	
\vspace{-5pt}
		\caption{Quantitative comparison for number of sentences $n_s$ for SenClu.}\label{tab:resns}
    \centering	%\begin{footnotesize}%\setlength\tabcolsep{2pt}    
		\begin{tabular} {| c | l | l  | l | }
			\hline			
			Dataset & \#Sen. $n_s$ & NMI  & PMI  \\ \hline
\multirow{3}{*}{20News} &  1 & 0.45\tiny{\text{$\pm$}0.02} & 0.76\tiny{\text{$\pm$}0.02} \\
&3 & 0.47\tiny{\text{$\pm$}.003} & 0.79\tiny{\text{$\pm$}.037} \\ 
 &  9 & 0.47\tiny{\text{$\pm$}0.0} & 0.79\tiny{\text{$\pm$}0.03} \\\hline
\multirow{3}{*}{Guten} &  1 & 0.19\tiny{\text{$\pm$}0.01} & 0.66\tiny{\text{$\pm$}0.07} \\
&3 & 0.2\tiny{\text{$\pm$}.003} & 0.67\tiny{\text{$\pm$}.03} \\ 
 &  9 & 0.21\tiny{\text{$\pm$}0.02} & 0.76\tiny{\text{$\pm$}0.04} \\\hline
\multirow{3}{*}{NYT} &  1 & 0.28\tiny{\text{$\pm$}0.02} & 0.75\tiny{\text{$\pm$}0.03} \\
&3 & 0.28\tiny{\text{$\pm$}.021} & 0.78\tiny{\text{$\pm$}.025} \\ 
 &  9 & 0.31\tiny{\text{$\pm$}0.03} & 0.79\tiny{\text{$\pm$}0.03} \\\hline
\multirow{3}{*}{Yelp} &  1 & - & 0.65\tiny{\text{$\pm$}0.04} \\
&3 & - & 0.62\tiny{\text{$\pm$}.007}\\ 
 &  9 & - & 0.65\tiny{\text{$\pm$}0.04} \\\hline
		\end{tabular}  
	%\end{footnotesize}	
	%\vspace{-6pt}
\end{table}

\begin{table}%[htp]	
\vspace{-5pt}
		\caption{Quantitative comparison for prior $\alpha$ for SenClu. }\label{tab:resalpha}
    \centering	%\begin{footnotesize}%\setlength\tabcolsep{2pt}    
		\begin{tabular} {| c | l | l  | l | }
			\hline			
			
   Dataset & $\alpha$ & NMI  & PMI  \\ \hline
\multirow{3}{*}{20News} & 0.25 & 0.43\tiny{\text{$\pm$}0.01} & 0.69\tiny{\text{$\pm$}0.03} \\
&2 & 0.47\tiny{\text{$\pm$}.003} & 0.79\tiny{\text{$\pm$}.037} \\ 
 & 8 & 0.49\tiny{\text{$\pm$}0.01} & 0.81\tiny{\text{$\pm$}0.02} \\\hline
\multirow{3}{*}{Guten} & 0.25 & 0.16\tiny{\text{$\pm$}0.01} & 0.73\tiny{\text{$\pm$}0.05} \\
&2 & 0.2\tiny{\text{$\pm$}.003} & 0.67\tiny{\text{$\pm$}.03} \\ 
 & 8 & 0.24\tiny{\text{$\pm$}0.01} & 0.78\tiny{\text{$\pm$}0.03} \\\hline
\multirow{3}{*}{NYT} & 0.25 & 0.24\tiny{\text{$\pm$}0.0} & 0.67\tiny{\text{$\pm$}0.0} \\
&2 & 0.28\tiny{\text{$\pm$}.021} & 0.78\tiny{\text{$\pm$}.025} \\ 
 & 8 & 0.33\tiny{\text{$\pm$}0.02} & 0.82\tiny{\text{$\pm$}0.03} \\\hline
\multirow{3}{*}{Yelp} & 0.25 & - & 0.58\tiny{\text{$\pm$}0.02} \\
&2 & - & 0.62\tiny{\text{$\pm$}.007}\\ 
 & 8 & - & 0.66\tiny{\text{$\pm$}0.02} \\\hline
		\end{tabular}  
	%\end{footnotesize}	
	%\vspace{-6pt}
\end{table}

\begin{table}%[htp]	
%\vspace{-5pt}
\vspace{15pt}%5
		\caption{Quantitative comparison between methods. Times are in minutes.}\label{tab:res}
    \centering
	\begin{footnotesize}
		\setlength\tabcolsep{2pt}    
		\begin{tabular} {| c | l | l  | l |l| }
			\hline			
			Dataset & Method & NMI  & PMI & Time  \\ \hline
            \multirow{4}{*}{20News} & BerTopic & 0.27 \scriptsize{$\pm$.011} & 0.2 \scriptsize{$\pm$.003} & 0.81 \scriptsize{$\pm$.012}\\
 & LDA & 0.24 \scriptsize{$\pm$.007} & 0.35 \scriptsize{$\pm$.002} & 0.31 \scriptsize{$\pm$.003}\\
 & TopClus & 0.38 \scriptsize{$\pm$.012} & 0.39 \scriptsize{$\pm$.016} & $>150$\\
 & \textbf{SenClu} & 0.47\tiny{\text{$\pm$}.003} & 0.79\tiny{\text{$\pm$}.037} & 2.26\tiny{\text{$\pm$}.031}\\ 
& \textbf{SenClu+FT-Topic}&0.48\tiny{\text{$\pm$} 0.005}&0.8\tiny{\text{$\pm$} 0.016}&25.36\tiny{\text{$\pm$} 0.269}\\ \hline 
\multirow{4}{*}{Guten} & BerTopic & 0.09 \scriptsize{$\pm$.0} & 0.44 \scriptsize{$\pm$.02} & 1.6 \scriptsize{$\pm$.122}\\
 & LDA & 0.25 \scriptsize{$\pm$.007} & 0.36 \scriptsize{$\pm$.022} & 0.83 \scriptsize{$\pm$.001}\\
 & TopClus & 0.24 \scriptsize{$\pm$.004} & 0.35 \scriptsize{$\pm$.014} & $>150$\\
 & \textbf{SenClu} & 0.2\tiny{\text{$\pm$}.003} & 0.67\tiny{\text{$\pm$}.03} & 5.62\tiny{\text{$\pm$}.37}\\ 
  & \textbf{SenClu+FT-Topic}&0.26\tiny{\text{$\pm$} 0.007}&0.74\tiny{\text{$\pm$} 0.035}&29.03\tiny{\text{$\pm$} 0.071}\\ \hline
\multirow{4}{*}{NYT} & BerTopic & 0.07 \scriptsize{$\pm$.009} & 0.2 \scriptsize{$\pm$.002} & 2.91 \scriptsize{$\pm$.05}\\
 & LDA & 0.21 \scriptsize{$\pm$.014} & 0.36 \scriptsize{$\pm$.008} & 1.24 \scriptsize{$\pm$.015}\\
 & TopClus & 0.25 \scriptsize{$\pm$.021} & 0.42 \scriptsize{$\pm$.009} & $>150$\\
 & \textbf{SenClu} & 0.28\tiny{\text{$\pm$}.021} & 0.78\tiny{\text{$\pm$}.025} & 6.47\tiny{\text{$\pm$}.158}\\ 
 & \textbf{SenClu+FT-Topic}&0.33\tiny{\text{$\pm$} 0.0}&0.85\tiny{\text{$\pm$} 0.0}&47.72\tiny{\text{$\pm$} 0.0}\\ \hline
\multirow{4}{*}{Yelp} & BerTopic & - & 0.15 \scriptsize{$\pm$.008} & 0.96 \scriptsize{$\pm$.115}\\
 & LDA & - & 0.32 \scriptsize{$\pm$.011} & 0.37 \scriptsize{$\pm$.006}\\
 & TopClus & - & 0.36 \scriptsize{$\pm$.008} & $>150$\\
 & \textbf{SenClu} & - & 0.62\tiny{\text{$\pm$}.007} & 2.91\tiny{\text{$\pm$}.045}\\ 
 & \textbf{SenClu+FT-Topic}&-&0.6\tiny{\text{$\pm$} 0.024}&21.74\tiny{\text{$\pm$} 1.052}\\ \hline
 
		\end{tabular}  
	\end{footnotesize}	
	%\vspace{-6pt}
\end{table}

\begin{table*}[ht]
   %\vspace{-6pt}
   \vspace{15pt}%5
   %\begin{adjustwidth}{-2.5cm}{}
		\caption{Top 7 words of topics by SenClu using FT-Topic and TopClus for first 15 of 50 topics }\label{tab:qua} % for 20Newsgroups dataset 
 \scriptsize
 \setlength{\tabcolsep}{0.5pt}
		\begin{tabular}{|l| l | l |}\hline
		To.	&20Newsgroups Dataset	& New York Times Dataset \\ \hline
  \multicolumn{3}{|c|}{Method: SenClu}\\ \hline
0 & \scriptsize{schism, papal, schismatic, excommunicated, excommunication, swinburne, pope} & \scriptsize{thai, curry, rice, korean, pad, panang, spicy}\\
1 & \scriptsize{homosexual, gay, homosexuality, sex, sexual, heterosexual, promiscuous} & \scriptsize{taco, guacamole, mexican, asada, tortilla, carne, salsa}\\
2 & \scriptsize{verse, scripture, sirach, lord, commandment, jesus, god} & \scriptsize{chinese, woo, sam, asian, dim, sum, chinatown}\\
3 & \scriptsize{bosnia, serb, bosnian, serbian, iraqi, irgun, saudi} & \scriptsize{pizza, crust, pepperoni, domino, hut, slice, oven}\\
4 & \scriptsize{limbaugh, rushdie, insult, cycnicism, racist, sarcasm, rebuttal} & \scriptsize{ramen, tonkotsu, shoyu, sora, tonkatsu, chashu, broth}\\
5 & \scriptsize{sin, jesus, god, christ, salvation, heaven, sinner} & \scriptsize{he, him, manager, his, she, her, apologized}\\
6 & \scriptsize{israel, israeli, arab, palestinian, gaza, zionist, palestine} & \scriptsize{thai, pad, thailand, cambodian, curry, papaya, panang}\\
7 & \scriptsize{widget, xlib, xterm, colormap, openwindows, window, sunos} & \scriptsize{noodle, handmade, shang, dumpling, wonton, broth, chow}\\
8 & \scriptsize{comic, shipping, marvel, wolverine, bagged, hulk, shatterstar} & \scriptsize{sandwich, vegan, healthy, kale, carnivore, abound, option}\\
9 & \scriptsize{president, stephanopoulos, myers, tax, republican, stimulus, deficit} & \scriptsize{burger, fry, cheeseburger, patty, bun, ring, pickle}\\
10 & \scriptsize{atheist, atheism, theist, belief, theism, existence, fallacy} & \scriptsize{ice, cream, cone, chocolate, yogurt, cupcake, shaved}\\
11 & \scriptsize{scorer, unassisted, mullen, nyr, det, nyi, pt} & \scriptsize{parking, park, carpet, bar, music, basement, room}\\
12 & \scriptsize{ax, max, pl, ei, jz, lk, ql} & \scriptsize{beer, biker, craft, brewery, tap, draft, brew}\\
13 & \scriptsize{constitution, amendment, libertarian, regulated, militia, infringed, tyranny} & \scriptsize{poisoning, roach, sick, vomiting, flu, enemy, dirty}\\
14 & \scriptsize{fbi, batf, koresh, compound, atf, raid, fire} & \scriptsize{sushi, roll, sashimi, nigiri, ayce, kama, maki}\\ \hline
 % 		\end{tabular}
 %  \vspace{-6pt}
	% 	\caption{Top 7 words of topics by SenClu  for first 15 topics }\label{tab:qua} % for 20Newsgroups dataset 
	% 	\vspace{-12pt}  
	% \end{table*}
 
 % 	\begin{table*}%[htp!]  		
 % \footnotesize
	% 	\begin{tabular}{|l| l | l |}\hline
	% 	Top	&20Newsgroups Dataset	& New York Times Dataset \\ \hline
\multicolumn{3}{|c|}{Method: TopClus} \\ \hline
0 & \scriptsize{please, thanks, thank, appreciate, sorry, appreciated, gladly} & \scriptsize{student, educator, grader, pupil, teenager, adolescent, school}\\
1 & \scriptsize{saint, biblical, messiah, missionary, apostle, church, evangelist} & \scriptsize{surname, mustache, syllable, corps, sob, nickname, forehead}\\
2 & \scriptsize{iranian, korean, hut, child, algeria, vegetable, lebanese} & \scriptsize{participation, involvement, effectiveness, supremacy, prowess, responsibility}\\%, productivity
3 & \scriptsize{considerable, tremendous, immense, plenty, countless, immensely, various} & \scriptsize{garage, dwelling, viaduct, hotel, residence, bungalow, building}\\
4 & \scriptsize{expression, phrase, symbol, terminology, prefix, meaning, coordinate} & \scriptsize{clit, lough, bros, kunst, mcc, quay, lund}\\
5 & \scriptsize{memoir, publication, hardcover, encyclopedia, bibliography, paperback} & \scriptsize{moth, taxa, una, imp, null, def, une}\\% columnist
6 & \scriptsize{anyone, somebody, anybody, someone, anything, everybody, something} & \scriptsize{many, everybody, anything, everyone, several, much, dozen}\\
7 & \scriptsize{individual, people, populace, human, being, inhabitant, peer} & \scriptsize{mister, iraqi, hussein, iraq, iranian, iran, kurdish}\\
8 & \scriptsize{disturbance, difficulty, complication, danger, annoyance, susceptible, problem} & \scriptsize{iraqi, iraq, baghdad, saddam, hussein, kuwait, iran}\\
9 & \scriptsize{beforehand, time, sooner, moment, waist, farther, halfway} & \scriptsize{dilemma, uncertainty, agitation, reality, dissatisfaction, implication, disagre.}\\%ement
10 & \scriptsize{upgrade, availability, replacement, sale, modification, repository, compatibility} & \scriptsize{nominate, terminate, establish, stimulate, locate, replace, protect}\\
11 & \scriptsize{buy, get, install, spend, sell, keep, build} & \scriptsize{withstand, hesitate, imagine, explain, apologize, happen, translate}\\
12 & \scriptsize{appropriated, reverted, wore, abolished, rescued, exercised, poured} & \scriptsize{forefront, accordance, extent, instance, way, precedence, behalf}\\
13 & \scriptsize{government, diplomat, fbi, ceo, parliament, officer, parliamentary} & \scriptsize{privy, continual, outstretched, purposely, systematically, unused, unfinished}\\
14 & \scriptsize{graduation, university, rural, upstairs, overseas, basement, undergraduate} & \scriptsize{cautious, goofy, arrogant, painful, cocky, hasty, risky}\\
 \hline
		\end{tabular}
%\end{adjustwidth}
		\vspace{-6pt}
	\end{table*}

%For diversity TopClus has a very small edge, but it lacks clearly in other measures such as PMI and NMI

\subsection{Qualitative Evaluation}
We showcase the top words from the initial 15 topics and compare these with the top-performing method from our evaluation and previous studies. We chose not to assign labels to the topics, mirroring the actual outcomes users encounter in real-world topic modeling scenarios. Nevertheless, to facilitate comprehension of the topics and the dataset, we have included the ground truth classes for the 20Newsgroups dataset in Table \ref{tab:top}. Our examination indicates that TopClus, similar to LDA, occasionally constructs topics using prevalent words that lack substantial contextual significance and should be omitted. For example, in the 20Newsgroups dataset, topics 0, 3, and 6 consist of commonly used but non-descriptive words, whereas topics 4, 8, and 9 prove difficult to classify. This problem, also evident in LDA, arises from the Bag of Words model. Despite these challenges, certain topics are distinctly understandable; for instance, topic 1 is closely associated with the ground truth category 'religion', and topic 11 aligns well with 'forsale', as detailed in Table \ref{tab:top} for ground truth classifications.

For SenClu with FT-Topic most topics are easy to interpret, e.g., Topic 10 deals with atheism, Topic 0 and 5 with religion, Topic 11 with hockey. Some are harder to assign to the given categories although they make sense, e.g., Topic 1 seems to be about sexuality, but it actually falls under the religion category, Topic 3 belongs to ``politics.mideast''. It does mention countries either from the Middle East or at least having cultural ties to the Middle East, but nothing about politics.  But it also contains a few topics, which make limited sense. For example, Topic 12 consists of tokens that likely stand for abbreviations of user-names, while Topic 14 requires some knowledge of US history, as it is about a famous and well-discussed event, e.g., the raid by the FBI on Koresh's compound during the Waco siege. %For people not familiar with the domain of the corpus some topics might appear too specific although still meaningful. For example, Topic 14 contains as first word `relegation', which might point towards competitive spots, but others are names of hockey players such as Whitmore, Berthiaume, and Fiset. These names though possibly well-known to hockey fans might not be known to the general public. Thus, overall, we deem that SenClu performs better than TopClus, since more topics are meaningful.

%\clearpage
 \begin{table*}[htp]%[b]{0.46\textwidth}
    % \vspace{-6pt}
    \vspace{15pt} %5
    \caption{Summarized Assessment of Methods}\label{tab:meth}  
    \centering
    %\begin{scriptsize}
        %\setlength\tabcolsep{1pt}
        \begin{tabular}{ l  c  c  c  c c}
            \toprule
         Method & Multiple topics  & Topics per doc & Speed & Topic quality & Method  \\	
          &  per doc? &  controllable?  & & complexity\\	
         \midrule
            LDA      & Y & Y & Fast & Medium & Low \\
            BerTopic & N & N & Fast & Medium-High & Low\\
            TopClus & Y  & N & Very Slow & High & Medium\\
            \textbf{SenClu}(ours) & Y & Y & Medium & High & Low\\                    
            \small{\textbf{SenClu+FT-Topic}(ours)} & Y & Y & Slow & High-Very Hi. & Low-Med.\\                    
         \bottomrule
        \end{tabular}
    %\end{scriptsize}  
   \vspace{-6pt}
\end{table*} 
\subsection{Overall Evaluation} \label{sec:over}
Table \ref{tab:meth} provides a high-level comparison of all methods, including both quantitative evaluations and the functionalities they offer. Despite being a very fast and conceptually elegant approach, LDA suffers in terms of topic quality, which is the most critical aspect of a topic model. As a result, it is less preferable compared to methods that rely on pretrained contextual embeddings, aligning with previous research findings \ci{men22,gro22}.

BerTopic, while also very fast, often falls short in topic quality and treats documents as having only one topic. This contradicts the fundamental idea of topic models that documents can encompass multiple topics. This limitation is particularly problematic for long, diverse texts where multiple topics are usually present.

TopClus produces high-quality topics but faces challenges in interpretability due to its reliance on training a neural network from scratch with multiple loss functions. Neural networks are notoriously difficult to interpret \ci{sch19c,mesk22,lon23}, and the computational overhead makes TopClus impractical for regular use. Additionally, it does not allow users to specify the desired number of topics per document, which can be a significant drawback. In contrast, LDA and our method include a hyperparameter ($\alpha$) that guides the algorithm to prefer few or many topics per document, offering a clear advantage.

In summary, SenClu achieves state-of-the-art topic quality within a reasonable timeframe and provides all functionalities desirable for users. Using a fine-tuned model for computing embeddings with our algorithm FT-Topic further enhances topic quality, although it significantly slows down the topic modeling training process. However, it does not affect inference times and adds complexity only in terms of parameter tuning, which users can avoid by relying on default settings.

\section{\uppercase{Related Work} }\label{sec:rel}  

\noindent\textbf{Early, discrete topic models:} Probabilistic Latent Semantic Analysis (PLSA)\ci{hof99} emerged in the previous millennium as an enhancement to Latent Semantic Analysis by incorporating discrete word representations like one-hot encodings. Latent Dirichlet Allocation (LDA)\ci{ble03} further developed this concept by adding priors with hyperparameters to sample from topic and word distributions, thereby generalizing PLSA. LDA has since been extensively modified and expanded. Unlike LDA, which calculates the generative probability of a word within a topic $p(w|t)$ based on the word's frequency in that topic, our approach posits that semantic similarity determines the probability $p(g|d)$ that a group of sentences, which is our unit of analysis rather than words, belongs to a topic. While many models focus on words, particularly using a bag-of-words approach, a few have explored single sentence assignments. For instance, \ci{gru07} assigns each sentence to a topic using a Markov chain to model transitions between topics after each sentence. Although technically different, this work also emphasizes a larger unit than word as a crucial unit of analysis, helping to prevent multiple nearby words from being assigned to different topics. \ci{bal16} introduces an additional "plate" of a sentence in an extension of LDA, where all words in a sentence are assigned to the same topic, though it does not address the underlying issue in LDA where frequency dictates likelihood. \ci{sch18} employs a strategy to identify keywords influencing the topic of surrounding words, which effectively results in chunks of text being assigned the same topic.

\smallskip

\noindent\textbf{Early topic models with continuous word representations:} Early works treated words as discrete entities through one-hot encodings. Following the success of static word vectors\ci{mik13} developed through neural networks, there was a shift towards utilizing continuous representations in topic modeling, with early examples including \ci{niu15,das15,mia16}. While using external knowledge to enhance topic models is well-known\ci{new11}, most efforts have focused on deriving vectors from the corpus intended for topic modeling. Neural topic models, which utilize deep learning networks for topic analysis, have become increasingly popular\ci{zha21s,dan22}, addressing challenges like accounting for correlated and structured topics\ci{xun17}, incorporating metadata, and accelerating inference\ci{sri17}. \ci{bia20} aims to merge a ProdLDA\ci{sri17} variant with document embeddings\ci{rei19}, simply incorporating document embeddings into the autoencoder input. Though improvements were noted over traditional LDA\ci{ble03} and ProdLDA\ci{sri17}, they were inconsistent across other models. 

\smallskip

\noindent\textbf{Topic models using LLM (encoders):}
\ci{hoy20} uses knowledge distillation by computing two word distributions: one from a standard topic model variant of LDA\ci{sri17} and another from a pretrained model like BERT, using these as a basis for training a student network to reconstruct documents. Our algorithm \emph{SenClu} aligns more closely with BERTopic\ci{gro22} and TopClus\ci{men22}, which also employ pre-trained contextualized embeddings (but without fine-tuning), coupled with some form of dimensionality reduction and clustering. Both argue the benefit of reducing dimensions, although it is important to consider that typical word embeddings\ci{rei19} are designed within a 300-800 dimensional space intended for large-scale data, suggesting that reduction might lead to loss of information. \ci{men22} posits that optimizing a dimensionality reduction layer specifically for clustering can enhance outcomes, placing some of the clustering burden on the reduction process itself, although this could also result in information loss, especially in smaller datasets often consisting of fewer than 100k documents. In our model, we avoid dimensionality reduction of pre-trained embeddings due to the potential for information loss and added complexity. In \ci{men22}, a word is represented as a product of the pre-trained embedding and an attention weight, with document embeddings summed from attention-weighted word embeddings. An objective is to optimally reconstruct these document sums by summing the topic embeddings of a document. Our model differs by implementing hard assignments, ensuring a sentence is assigned to only one topic, which seems more intuitive from a human topic modeling perspective and reduces computational demands. \ci{men22} also noted potential issues with soft assignments, attempting to address these by squaring the topic-word distribution, which accentuates differences between the most and second most likely topics, though this approach is somewhat arbitrary conceputally. The training requires managing three distinct losses, each needing to be weighted, making it more complex and computationally more intensive than our method. BERTopic\ci{gro22} processes entire document embeddings through contextual word vectors\ci{rei19} and clusters them using a density-based clustering technique, namely HDBScan, which overlooks the potential for documents to encompass multiple topics. It is therefore more similar to document clustering than classical topic modeling. Unlike BERTopic, our approach and other models accommodate the possibility of multiple topics per document, making our clustering approach akin to K-Means but with a more detailed computation of topic-sentence probabilities, thus rendering our clustering process more sophisticated. Prior to using continuous representations, i.e., before our work and \ci{gro22}, various methods were developed to integrate document clustering and topic modeling in ways differing from our approach, e.g.,\ci{xie13}.

\smallskip

\noindent\textbf{Pretrained language models:} Early efforts to learn word vectors, forms of distributed representation, date back to the early 2000s\ci{ben00} and gained prominence about a decade later through a simple neural network architecture that produced static word vectors from large corpora, enabling arithmetic operations on words\ci{mik13}. Contextual word embeddings\ci{dev18} succeeded static vectors using more complex transformer architectures, allowing for the derivation of vectors based on the context of a word, e.g., a word and its surrounding text. Since their introduction, numerous enhancements have been suggested, including models tailored for embedding sentences\ci{rei19}, improving robustness\ci{liu19}, and enhancing performance\ci{san19}. While employing a sentence embedder is the obvious choice for a bag of sentence models, other models could also be utilized and could offer benefits, e.g., for faster inference, words within a sentence might be aggregated using a fast version of BERT\ci{san19}.

\noindent\textbf{Topic modeling using LLMs: }
Topic modeling can also be performed using large language models \cite{pha24,sch25hi}. This is simple for the user and also enables easy customization. For example,  \cite{sch25hi} investigated deriving hierarchical topic models using ChatGPT, which are difficult to obtain in high quality using other approaches. However, the use of large language models also has downsides. For once, the context window is limited making it impossible to conduct topic modeling naively on very large document collections. That is, special processes like iteratively summarizing documents would be needed. Our work is free from this limitation. Also, computation time of large language models might scale quadratically for large contexts depending on the attention mechanism. Whereas our method is linear and significantly faster. Our work is also more explainable as it maintains sentence to topic assignments for each document, which can easily be investigated by a user. Explainability is an essential problem in AI research \cite{sch24genx,mesk22,sch19c}. Also the impact of biases on outcomes is larger for methods that only rely on LLMs. That is, we only use LLM (encoder) for similarity computation, but each sentence still becomes assigned to a topic. While methods relying only on LLMs (like \cite{pha24,sch25hi}) do not provide such guarantees, i.e., the LLM might neglect some information completely, while hallucinating non-existing topic data. Furthermore, prompt sensitivity can be a problem \cite{sch24fou}.
Thus, we believe that both methods will prove valuable in the future.

\smallskip

\noindent\textbf{Topic labeling }  addresses the challenge of identifying suitable descriptions for detected topics or paragraphs. Typically, external resources like Wikipedia are utilized to perform this task\ci{lau11}. Our approach, like those of \ci{sch18} and \ci{gro22}, relies solely on corpus-inherent knowledge to pinpoint topic words. At a high level, all methods follow a similar strategy, weighing words based on measures accounting for the frequency of a term and its distribution across topics, as done in classic metrics like term frequency and inverse document frequency (TF-IDF)\ci{has14}.
%\noindent{Document clustering} 

\smallskip

\noindent\textbf{Fine-Tuning and topic modeling } Fine-tuning of foundation models towards topic models has only be performed for diffusion models. However, in contrast to our work they relied on human-labelled data, where we generate data in an unsupervised manner, which is a major plus\cite{xu23d}. While for multimodal topic modeling, fine-tuning has been stated as a research gap to be studied \cite{pra23}.	

\section{\uppercase{Discussion and Future Work}}
Contextual word embeddings generated through transformers have advanced the state-of-the-art in natural language processing (NLP). In our research, we showed how to fine-tune LLM-based encoders used within topic models to improve outcomes. We did not tune all parameters and we firmly believe that, e.g., by enhancing the training data for fine-tuning (e.g., by using more negative samples and more training epochs) and performing dataset specific tuning, performance can be further enhanced. 
Furthermore, our topic model SenClu relies on a bag of sentence model  that could benefit from well-established concepts within the bag of words framework, such as the relevance of sentence proximity to topic coherence. Integrating these insights could enhance topic model accuracy but at the cost of increased algorithmic complexity. Complex and slow algorithms present significant challenges, particularly for users with less powerful computing resources who need to experiment with various hyperparameter settings. While our current model operates swiftly, we anticipate further optimizations in future developments. For instance, implementing an expectation-maximization (EM) step with a subset of documents could expedite convergence. Enhancements in sentence tokenization and word embeddings could also augment our model's performance.

Our approach builds upon the foundational aspect model, presenting it from a fresh perspective. Alternatively, viewing it through the lens of k-Means clustering, which also employs expectation maximization to calculate cluster centers by averaging all assigned points, offers another angle. Unlike k-Means that deals with individual points, our model works with nested sets of points, demanding a hierarchical approach to clustering. Such an arrangement necessitates modifications like an "annealing" process to ensure the generation of high-quality topics.
Our work relies on a heuristic for obtaining data for fine-tuning: Nearby sentences in a document are more likely to share the same topic than random sentences from other documents. This heuristic can introduce errors. It might be further improved by applying an iterative data-labeling process as common in semi-supervised learning.
Our work performs hard assignments of sentences to topics. While this notion has computational advantages and implies less mental load for users, it also limits the flexibility of topic assignments compared to soft assignments.
Our work is not ensured to reach  global maxima. Future work might also investigate running our algorithm multiple times to avoid poor solutions - a strategy common in machine learning, e.g., for k-Means.
Most topic modeling methods come with a number of parameters, and our work is no exception. While some relate to user preferences, others are algorithm specific. For our algorithm parameter sensitivity is low, meaning that it should be possible to rely on provided defaults. Still, any parameter is a burden on the user. Our algorithm is linear in the number of terms, i.e., documents times (max) document length, i.e., its complexity is $O(|D|\max_{d \in D} |d|)$. This is asymptotically optimal as each term must be used once in the computation. However, the hidden constants such as number of epochs can be significant in practice. Both the E- and the M-step are highly parallelizable.
Future work might also investigate integrating using LLMs and our approach. For example, LLMs might be useful to generate a textual description of topics by summarizing sentences assigned to a topic. Such a description is more interpretable for users. The overarching idea to use LLMs to explain algorithm outcomes is well-studied, e.g., \cite{bok25}.

\section{\uppercase{Conclusions} }\label{sec:lim}
Topic modeling remains a complex area within natural language processing (NLP). Traditional topic models, still widely used, are outdated and fraught with significant limitations. Our research has contributed to the emerging field leveraging LLM-based encoders by suggesting how to leverage them in an unsupervised manner using a method called FT-Topic. Furthermore, our topic model SenClu effectively used contextual word embeddings, integrating them into a novel topic modeling approach that addresses these deficiencies. This new model not only drastically reduces computation times and enables the extraction of multiple topics from a single document but also enhances performance across various measures, including applications in downstream tasks. By incorporating elements such as a Bag-of-Sentences structure, hard assignments strategies, and simulated annealing in our inference methods, we've significantly improved topic model outputs. While our experimental results are promising, we acknowledge that there is room for further enhancement. We invite other researchers to explore these possibilities and contribute to advancing this challenging field.

Topic modeling is challenging.  While many other NLP tasks have rapidly advanced in the last years, wildly used topic models still date back decades despite striking weaknesses. Our work has shown how to utilize external knowledge in the form of contextual word embeddings in an efficient manner to improve on prior topic models. It not only overcomes major shortcomings of prior works such as extremely long computation times or the inability to extract multiple topics per document, but it also improves on a variety of other measures such as downstream tasks. To do so, we utilize and introduce a novel topic model including inference mechanisms based on multiple ideas such as Bag-of-Sentences, hard assignments, and simulated annealing. Despite our promising experimental evaluation, we have elaborated in our discussion that  further improvements might be possible and encourage fellow researchers to engage in this challenge.
%\noindent \textbf{Ethical Statement:} There are no ethical issues.

\section{Declarations}

- Competing Interests\\
Not applicable
- Funding Information\\
Not applicable
- Author contribution\\
Not applicable
- Data Availability Statement\\
Not applicable
- Research Involving Human and /or Animals\\
Not applicable
- Informed Consent\\
Not applicable

\bibliographystyle{apalike}
{\small
\bibliography{refs}}

\end{document}